\documentclass[runningheads]{llncs}

\usepackage[mobile]{eccv}

\usepackage{eccvabbrv}
\usepackage{bm}
\usepackage{wrapfig}
\usepackage{graphicx}
\usepackage{booktabs}
\usepackage{multicol}
\usepackage{multirow}
\usepackage{soul}
\usepackage[accsupp]{axessibility}  

\usepackage[pagebackref,breaklinks,colorlinks,citecolor=eccvblue]{hyperref}

\usepackage{orcidlink}

\begin{document}

\title{URS-NeRF: Unordered  Rolling  Shutter Bundle Adjustment  for Neural Radiance Fields } 

\titlerunning{ }

\author{Bo Xu \inst{1, 2}, 
Ziao Liu \inst{1}, 
Mengqi Guo \inst{2}, 
Jiancheng Li\inst{1},
Gim Hee Lee\inst{2}}

\authorrunning{ }

\institute{Wuhan University\and
National University of Singapore}
\maketitle

\begin{abstract}
We propose a novel rolling shutter bundle adjustment method for neural radiance fields (NeRF), which utilizes the unordered rolling shutter (RS) images to obtain the implicit 3D representation. Existing NeRF methods suffer from low-quality images and inaccurate initial camera poses due to the RS effect in the image, whereas, the previous method that incorporates the RS into NeRF requires strict sequential data input, limiting its widespread applicability. In constant, our method recovers the physical formation of RS images by estimating camera poses and velocities, thereby removing the input constraints on sequential data. Moreover, we adopt a coarse-to-fine training strategy, in which the  RS epipolar constraints of the pairwise frames in the scene graph are used to detect the camera poses that fall into local minima.  The poses detected as outliers are corrected by the interpolation method with neighboring poses. The experimental results validate the effectiveness of our method over state-of-the-art works and demonstrate that the reconstruction of 3D representations is not constrained by the requirement of video sequence input.

\end{abstract}

\section{Introduction}

{NeRF  \cite{mildenhall2021nerf} has recently emerged as a ground-breaking implicit 3D representation that provides new perspectives for computer vision and graphics.}
{The prerequisites to learn good representation of a 3D scene with NeRF are high-quality images and accurate camera poses.}
{However, it can be challenging to acquire such high-quality images and accurate camera poses from the commonly used RS cameras due to rolling shutter distortions caused by sequential scanning time of each row or column of the images taken from a moving camera \cite{meingast2005geometric}. Neglecting the distortions in images and inaccurate pose estimations due to the rolling shutter effect is detrimental to learning 3D representations with NeRF. Nonetheless, since NeRF has become a de-facto approach for learning 3D representations and RS cameras are widely used in many consumer products such as mobile phones due to inexpensive cost, low energy consumption and high-frame rates, it is therefore imperative to propose a framework for NeRF with rolling shutter images.       
}

{Many works \cite{zhuang2017rolling, liao2023revisiting, saurer2016sparse, hedborg2012rolling, patron2015spline, lao2018robustified} propose the adaptation of bundle adjustment (BA) with the RS camera model to improve RS camera pose estimation and remove rolling shutter distortions. The inputs to these methods are either sequential video frames or unordered images.} 
{Methods based on sequential video frames \cite{hedborg2012rolling, patron2015spline} can leverage the connection between consecutive frames to improve motion estimation results. 
For example, the use of smooth motion assumptions between consecutive frames and the spline-based trajectory model provides an effective way of reconstructing the 3D scenes with rolling shutter images. However, these constraints do not work for the more general unordered image setting and are susceptible to loop-closure errors when revisiting previously visited scenes. To circumvent the lack of consecutive frame constraints, the linear and angular velocities of the unordered camera images 
can be estimated under the assumption of constant motion during the exposure time. Although this weaker constraint would result in decreased accuracy, it can be easily mitigated with bundle adjustment. 
}

\begin{figure}[t!]
  \centering
  \includegraphics[width=\textwidth]{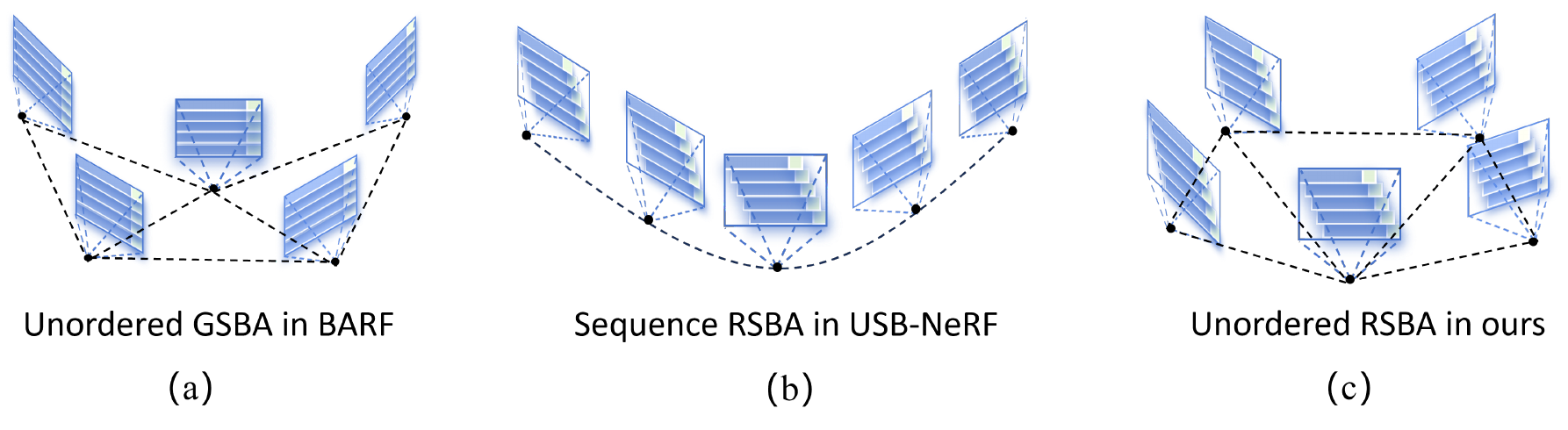}
  \vspace{-2em}
  \caption{
  The illustration of different NeRF BA settings. (a) BARF \cite{lin2021barf} works on unordered global shutter (GS) images, but is unsuitable for images with rolling shutter distortions. (b) USB-NeRF \cite{li2023usb} requires the input to be strictly sequential video frames, which lacks the generality in BA. (c) Our method works on unordered images with rolling shutter distortion. }
  \label{fig:abstract_figure}
  \vspace{-2.5em}
\end{figure}

NeRF demonstrates the impressive capability
of encoding the implicit scene representation and rendering high-quality images at novel views with only a small set of coordinate-based MLPs. 
{This leads to USB-NeRF \cite{li2023usb} in introducing} rolling shutter model into BARF \cite{lin2021barf} to learn the 3D representation and recover the camera motion trajectory simultaneously. However, the spline-based trajectory model used in USB-NeRF requires 
{the input to be strictly from sequential video frames, and therefore lacks the generality of learning the 3D scene from unordered images.} 
{In view of this limitation, we propose rolling shutter bundle adjustment for neural radiance fields using unordered images in this paper.} 
The differences between the various methods are demonstrated in 
Fig.~\ref{fig:abstract_figure}. A rolling shutter model that estimates the camera pose, linear, and angular velocities to recover the camera motion corresponding to each row of the rolling shutter image is introduced into our method. 
The 3D representation and the camera motion are {then} trained by maximizing the photometric consistency between the rendered and captured rolling shutter images. To achieve stable optimization, we adopt a coarse-to-fine optimization strategy and utilize rolling shutter epipolar geometry constraints of the pairwise frames in the scene graph to detect the camera poses that are sub-optimal. The poses detected as outliers are corrected by the interpolation method with neighboring poses.  Our method allows for the synthesis of high-quality GS images from novel view and the generation of datasets with varying degrees of rolling shutter effects, which is significant for the rolling shutter research community, \eg rolling shutter bundle adjustment and simultaneous localization and mapping (SLAM). 

{
Our \textbf{main contributions} are summarized as follows:
\begin{itemize}
    \item We propose URS-NeRF which puts the RS camera model together with bundle adjustment and Tri-MipRF \cite{hu2023Tri-MipRF} for rolling shutter effect removal, novel view image synthesis, and RS camera pose and velocity estimations from unordered rolling shutter images.
    \item We introduce a coarse-to-fine strategy to prevent the bundle adjustment with rolling shutter motion parameters and NeRF from getting trapped in local minima. We further suggest a strategy to check for the erroneous pose in the bundle adjustment using the rolling shutter epipolar constraints of the pairwise frames in the scene graph.    
    \item Extensive experimental evaluations are conducted on both synthetic and real datasets to evaluate the performance of our method. The experimental results demonstrate that our method achieves superior performance in terms of rolling shutter effect removal, novel view image synthesis, and camera motion estimation.
\end{itemize}
}

\vspace{-2mm}
\section{Related Work}
\vspace{-2mm}
\subsubsection{Rolling Shutter Effect Correction.} The 3D reconstruction with RS images has been widely studied. Depending on the type of input data (e.g., video sequence or unordered images) \cite{liao2023revisiting}, the RSBA methods employ different models to formulate the camera motion corresponding to different scanlines within the exposure time of the image. Im et al. \cite{im2018accurate} make use of an RS video to solve RSSfM and present a small motion interpolation-based RSBA algorithm applicable to compensate for the rolling shutter effect. To model more complex motion of the camera, Patron et al. \cite{patron2015spline} propose a spline-based trajectory model to better reformulate the RS camera motion between consecutive frames. Zhuang et al. develop a 9-point algorithm to estimate the relative pose from two consecutive RS images. The high-quality GS images can also be recovered with the relative poses. Despite the promising results achieved by using video-based RS methods,  the overly restrictive constraints on input images still affect the applications, especially in case of the classical SfM pipeline. Albl et al. \cite{albl2016degeneracies} address the unordered RSSfM problem and point out the planar degeneracy configuration of RSSfM. Liao et al. \cite{liao2023revisiting} proposes a normalization and covariance standardization weighting RSBA method that can be used to recover the camera poses with independent RS inputs.  Unlike the introduction of velocity parameters or imposition of continuous time and motion through pose interpolation, a local differential RS constraint is proposed by Lao et al.  \cite{lao2021solving} to deal with RS effects in SfM. With the development of deep learning,  There are many methods proposed for RS effect correction with the network. Rengarajan et al. \cite{rengarajan2017unrolling} propose a convolutional neural network (CNN) to estimate the row-wise camera motion from a single RS image. Fan et al.  \cite{fan2021sunet} recover the global shutter image from two consecutive images with unrolling shutter networks. Furthermore, Fan et al. \cite{fan2022context} present a refined scheme under which the bilateral motion field recovered from two RS frames is used to produce high-fidelity GS video frames at arbitrary times. However, these methods usually require the use of a large dataset to complete training, and the generalization performance is constrained on the images of different characteristics, as verified in \cite{li2023usb}. On the contrary, our approach does not require any pre-trained models, thus demonstrating superior generalization capabilities.

\vspace{-2mm}
\subsubsection{Neural Radiance Fields.} Recently, NeRF \cite{mildenhall2021nerf} has attracted widespread attention due to its impressive capability to represent 3D scenes. Plenty of extensions are proposed for better performance in practice. To complete the training of NeRF with inaccurate or unknown cameras, a series of improvement algorithms \cite{fu2023cbarf, lin2021barf, chen2023dbarf, song2023sc} have been proposed to optimize the network and camera poses simultaneously. Meanwhile, there are a lot of works focusing on how to improve the rendering quality \cite{barron2021mip, barron2022mip, barron2023zip} and training speed \cite{mueller2022instant, yu2021plenoctrees, chen2022tensorf}.  
\cite{muller2022instant} address these concerns with multi-resolution hash encoding, which achieves instant reconstruction in around five minutes and rendering in real-time.  \cite{hu2023Tri-MipRF} propose a Tri-Mip encoding into NeRF, which achieves high-fidelity anti-aliased renderings and efficient reconstruction.  These works need both high-quality GS images and corresponding accurate camera poses, which is not suitable for the RS task.  Li et al. \cite{li2023usb} propose unrolling shutter bundle adjusted neural radiance fields, in which the motion trajectory of the RS video sequence is parameterized with the cubic B-Spline interpolation method. However, the method based on BARF requires a lengthy training time.
Moreover, there is a limit on its applicability due to the highly complex constraints on the input. 
In contrast, our method does not have these limitations and works on the unordered rolling shutter images.

\vspace{-2mm}
\section{Notations and Preliminaries}
\subsection{Bundle Adjusting Neural Radiance Fields}

{BARF \cite{lin2021barf} is the first work to present bundle adjusting NeRF.}
Given the camera view with 
pose $\mathbf{T}^w_c=\left(\mathbf{R}^w_c, \mathbf{t}^w_c\right)$, 
a simple neural network such as MLPs is used to 
output the color $\mathbf{c}=\left(r,g,b\right)$ and volume density $\sigma$ for each location $\mathbf{x}$ and camera view direction $\mathbf{d}$ in a 3D scene. 
{Using $N$} 3D points $\mathbf{p}^c$ sampled along a ray in the camera frame $\mathbf{r}\left(t\right) = \mathbf{o} + t \mathbf{d}$, {the} 3D position $\mathbf{p}^w$ in the world frame can be computed by:
$\mathbf{p}^w=\mathbf{T}^w_c \mathbf{p}^c$.
The color and volume density of the sampled point $\mathbf{p}^w$ can {then} be obtained as: $\left(\mathbf{c}, \sigma\right)=F\left(\mathbf{x}^w, \mathbf{d}^w\right)$, where $\mathbf{d}^w=\mathbf{R}^w_c \mathbf{d}^c$ is the viewing direction of the ray in the world frame. 
{After getting} the point color $\mathbf{c}_n$ and volume density $\sigma_n$ of all the $N$ points, the per-pixel RGB $\mathbf{c}\left(\mathbf{r}\right)$ value can be computed, \ie: 
\begin{equation}
c\left(\mathbf{r}\right) = \sum^N_{i=1} T_i\left(1-\exp{\left(-\sigma_i\delta_i\right)}\right)\mathbf{c}_i, \quad T_i=\exp{(-\sum_{j=1}^{i-1}\sigma_j\delta_j)},
\end{equation}
where $\delta_i$ indicates the distance between $i^{th}$ sample and $\left(i+1\right)^{th}$ sample, and $T_i$ is the accumulated transmittance along the ray $\mathbf{r}$ from camera center to $i^{th}$ 3D point.Finally, the image synthesis process 
{is given by:}
\begin{equation}
    \hat{\mathbf{I}} = \mathcal{C}\left(\mathcal{F}\left(\omega\left(\mathbf{p}^{c_1}, \mathbf{T}\right); \mathbf{\Theta}\right), \ldots, \mathcal{F}\left(\omega\left(\mathbf{p}^{c_k}, \mathbf{T}\right);\mathbf{\Theta} \right)\right),
\end{equation}
where $\mathcal{C}\left(\cdot\right)$ denotes the ray composition function, $\mathcal{F}\left(\cdot\right)$ indicates the NeRF network. $\omega\left(\cdot\right)$ is the rigid transformation which projects the point $\mathbf{p}$ from the camera frame to the world frame by the camera pose $\mathbf{T}$, and $\Theta$ indicates the network parameters.

Since the whole pipeline is differentiable, camera poses $\mathbf{T}$ and MLP network can be {jointly} optimized by supervising the rendering output and RGB image with the L2 distance:
\begin{equation}\label{equ: loss_fuction}
\mathcal{L}_{rgb}=\sum_{i}^{N}\sum_{\mathbf{d}} ||\hat{\mathbf{I}}_i - \mathbf{I}_i\left(\mathbf{d}\right)||,
\end{equation} 
where N is the total number of images in the training dataset.

\subsection{Tri-Mip encoding}
To achieve both high-fidelity anti-aliased renderings and efficient reconstruction, Tri-Mip encoding is introduced into the implicit neural radiance field \cite{hu2023Tri-MipRF}. 
{Instead of} performing ray casting that ignores the area of the pixel in NeRF, the rendered pixels are formulated as a disc on the image plane. The radius of the disc can be 
{computed} by $\dot{r} = \sqrt{\mathrm{\Delta} x \cdot \mathrm{\Delta} y / \pi}$, where $\mathrm{\Delta} x$ and $\mathrm{\Delta} y$ are the width and height of the pixel in world coordinates. For each pixel, a cone casting is performed from the camera projection center $\mathbf{o}$ along the camera view direction $\mathbf{d}$. 
The cone are {then} sampled with a set of spheres $S\left(\mathbf{x}, \mathbf{r}\right)$ that are inscribed with the cone, {where} the center $\mathbf{x}$ and radius $\mathbf{r}$ of the sphere can be written as:
\begin{equation}
\mathbf{x} = \mathbf{o} + t \mathbf{d}, \quad
\mathbf{r} = \frac{||\mathbf{x}-\mathbf{o}||_2 \cdot f \dot{r}}{||\mathbf{d}||_2 \cdot \sqrt{\left(\sqrt{||\mathbf{d}||_2^2-f^2}-\dot{r}\right)^2 + f^2}},
\end{equation}
where $f$ is the focal length. 
Furthermore, the spheres $S\left(\mathbf{x}, \mathbf{r}\right)$ are represented as feature vectors $\mathbf{f}$ by the Tri-Mip encoding that is parameterized by three trainable mipmaps $\mathcal{M}$:
\begin{equation}
\mathbf{f} = \mathrm{Tri\mbox{-}Mip}\left(\mathbf{x}, \mathbf{r}; \mathcal{M} \right), \quad
\mathcal{M} = \{\mathcal{M}_{XY}, \mathcal{M}_{XZ}, \mathcal{M}_{YZ}\}.
\end{equation}
To make reconstructed scene coherent at different distance, the base level $\mathcal{M}^{L_0}$ is randomly initialized and other levels $\left(\mathcal{M}^{L_i}, i=1, 2, \cdots, N\right)$ are derived from the previous level $\mathcal{M}^{L_{i-1}}$ by downscaling $2\times$ along the height and width {for each mipmap.}

\vspace{-1mm}
\section{Our Method}
\begin{figure}[t]
  \centering
  \vspace{-1mm}
  \includegraphics[width=0.9\textwidth]{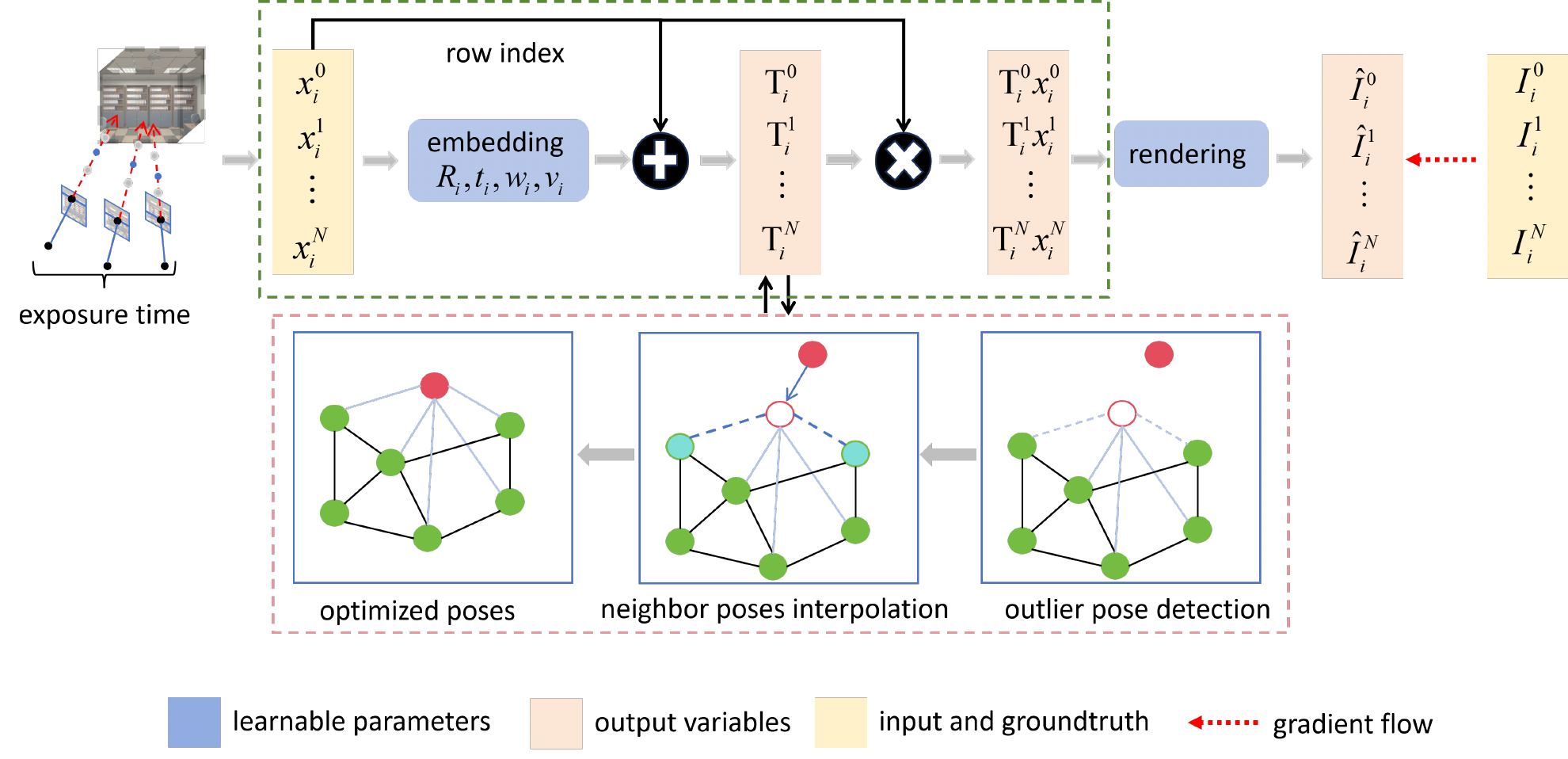}
  \vspace{-3mm}
  \caption{Overall pipeline of our proposed framework. We adopt a coarse-to-fine strategy to train rolling shutter images, and the scene graph is used to detect and correct the estimated poses that belong to outliers.  {Refer to the text for more details.}}
  \label{fig:framework}
  \vspace{-4mm}
\end{figure}
{
\vspace{-1mm}
\noindent \textbf{Overview.}
Figure~\ref{fig:framework} shows the illustration of our framework. Since the image captured with a RS camera is exposed row-by-row, the pose corresponding to each scanline is different. Based on the query coordinates $\{\mathbf{x}{^j}\}_{j=0}^{ N}$ in each rolling shutter image, our model constructs line-wise transformations $\{\mathbf{T}^j_i\}_{i=0, j=0}^{M, N}$ with the frame-dependent parameters $\mathbf{R}_i, \mathbf{t}_i, \bm \omega_i, \mathbf{v}_i$ and the row index of the sampled ray (\cf Sec.~\ref{sec:rollingshuttermodel}). Subsequently, the sampled rays are transformed from the query coordinates into the global coordinates. Finally, the color of each pixel can be rendered to get the rendered image $\{\mathbf{\hat{I}}{^j}\}_{j=0}^{ N}$, which we use to minimize the photometric error with the given image $\{\mathbf{I}{^j}\}_{j=0}^{ N}$ in our bundle adjustment formulation (\cf Sec.~\ref{sec: coarse to fine bundle adjustment}). To detect sub-optimal camera poses, a scene graph is constructed according to the number of the matched keypoints. The poses detected as outliers are corrected by the interpolation method with neighboring poses (\cf Sec.\ref{sec:Erroneous pose detection}).
}

\vspace{-1mm}
\subsection{Rolling Shutter Camera Model}\label{sec:rollingshuttermodel}

\begin{figure}[t!]
        \centering
	\vspace{-1mm}
	   \includegraphics[width=0.65\textwidth]{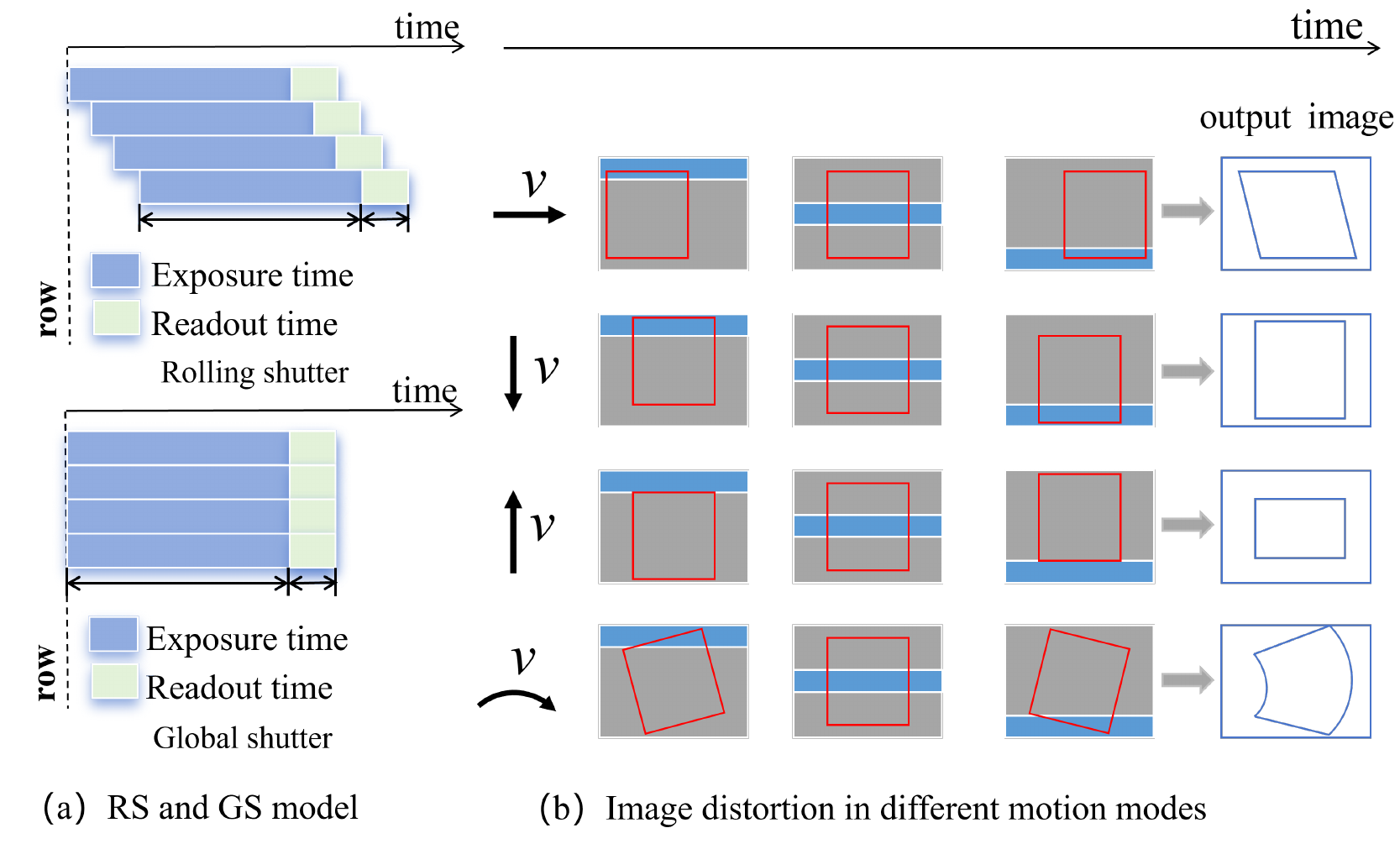}
    \vspace{-1em}
	\caption{(a) Image formation models of a RS camera (top) and a GS camera  (bottom). (b) Final image shapes of different motion modes for RS camera. It demonstrates that each row of a rolling shutter image is captured at different timestamps, and would thus lead to different image distortions if the image is captured by a moving camera.}
	\label{fig:rolling_shutter_effect}
	\vspace{-5mm}
\end{figure}

\vspace{-1mm}
{In contrast to} GS cameras {where the whole image is captured simultaneously}, each scanline of the RS camera is captured at different timestamps. {Consequently, significant rolling shutter distortions appear in 
the image when the camera undergoes large motion, as can be seen in Fig~\ref{fig:rolling_shutter_effect}.}
{It is, therefore, difficult to obtain accurate prior RS camera poses using COLMAP \cite{schoenberger2016sfm} for implicit neural radiance fields.}
{Since} the rows of an image 
{are} not taken at the same time, it is necessary to find the camera pose $\mathbf{T}^j\left(t\right)$ as a function of time $t$ to model the RS camera. We assume that the time $t$ when a pixel is read out is linearly related to the vertical y-coordinate of the image. 
{Furthermore,} the camera motion is assumed {to be} constant during frame capture, which usually is well-satisfied for RS cameras. We {follow 
\cite{albl2016degeneracies, dai2016rolling,lao2021solving} to} model the instantaneous-motion as: 
\begin{equation}
\mathbf{R}^j\left(u_i\right) = \left(\mathbf{I} + [\bm{\omega}^j]_{\times} {u_i}\right)\mathbf{R}^j_0, \quad 
\mathbf{t}^j\left(l_i\right) = \mathbf{t}^j_0 + \mathbf{v}^ju_i,
\end{equation}
where $\mathbf{R}^j\left(u_i\right) \in \mathbf{SO} \left(3\right)$ and $\mathbf{t}^j\left(u_i\right) \in \mathbb{R}^3$ define the camera rotation and translation when the row index $u_i$ is acquired, respectively. $\left[\bm{\omega}^j\right]_{\times}$ represents the skew-symmetric {cross-product} matrix of vector $\bm{\omega}^j$. $\mathbf{R}^j_0$ and $\mathbf{t}_0^j$ are the rotation and translation matrix when the first row is observed. $\mathbf{v}^j = \left[v^j_x, v^j_y, v^j_z\right]^{\top}$ is the linear velocity vector and $\bm{\omega}^j = \left[\omega^j_x, \omega^j_y, \omega^j_z\right]^{\top}$ is the angular velocity vector. 
Therefore, the rolling shutter image synthesis process can be written as:
\begin{equation}\label{equ: rolling shutter nerf}
    \hat{\mathbf{I}}^r = \mathcal{C}\left(\mathcal{F}\left(\mathbf{f}\left(\omega \left(\mathbf{p}^{c_1}, \mathbf{T}, \bm{\omega}, \mathbf{v}\right)\right); \mathbf{\Theta}\right), \ldots, \mathcal{F}\left(\mathbf{f}\left(\omega\left(\mathbf{p}^{c_k}, \mathbf{T}, \bm{\omega}, \mathbf{v}\right)\right);\mathbf{\Theta} \right)\right)
\end{equation} 
where $\mathbf{f}\left(\cdot\right)$ is the Tri-Mip encoding. By supervising the synthesis output and RGB image with Eq. \ref{equ: loss_fuction}, we can train the MLP $\mathbf{\Theta}$ and mipmaps $\mathcal{M}$ representing the scene obtained by a GS camera 
{with} rolling shutter images, and then synthesize novel view global shutter images. 
{It is worth noting that when $\bm{\omega}$ and $\mathbf{v}$ are set to zero, the training of the network degenerates to using global shutter images.}

\vspace{-1mm}
\subsection{Coarse-to-fine Bundle Adjustment} 
\label{sec: coarse to fine bundle adjustment}

\vspace{-1mm}
To train the network with Tri-Mip encoding 
{that accounts for} the rolling shutter effect, we add $\mathcal{K} \times 6$ learnable pose embedding, $\mathcal{K} \times 3$ learnable linear velocity embedding and $\mathcal{K} \times 3$ learnable angular velocity embedding in the Tri-MipRF \cite{hu2023Tri-MipRF}. 
{Subsequently,} the gradients of $\mathcal{L}_{rgb}$ with respect to camera pose $\mathbf{T}_i$, linear velocity vector $\mathbf{v}^i$ and angular velocity vector $\bm{\omega}^i$ are derived {from Eq.~\ref{equ: rolling shutter nerf}} as: 
\begin{subequations}
\begin{equation} \label{equ: partial respective pose}
\frac{\partial \mathcal{L}_{rgb}}{\partial \mathbf{T}_i} = \sum_i^N \sum_\mathbf{d}\sum_k^K \frac{\partial \mathcal{C}}{\partial \mathcal{F}_k} \cdot \frac{\partial \mathcal{F}_k}{\partial \mathbf{f}^k_i} \cdot \frac{\partial \mathbf{f}^k_i}{\partial \mathbf{T}_i}, 
\end{equation}
\begin{equation} \label{equ: partial respect to v}
\frac{\partial \mathcal{L}_{rgb}}{\partial \mathbf{v}_i} = \sum_i^N \sum_\mathbf{d}\sum_k^K \frac{\partial \mathcal{C}}{\partial \mathcal{F}_k} \cdot \frac{\partial \mathcal{F}_k}{\partial \mathbf{f}^k_i} \cdot \frac{\partial \mathbf{f}^k_i}{\partial \mathbf{v}_i}, 
\end{equation}
\begin{equation} \label{equ: partial respect to av}
\frac{\partial \mathcal{L}_{rgb}}{\partial \bm{\omega}_i} = \sum_i^N \sum_\mathbf{d}\sum_k^K \frac{\partial \mathcal{C}}{\partial \mathcal{F}_k} \cdot \frac{\partial \mathcal{F}_k}{\partial \mathbf{f}^k_i} \cdot \frac{\partial \mathbf{f}^k_i}{\partial \bm{\omega}_i}. 
\end{equation}
\end{subequations}

From the perspective of chain rule differentiation, introducing the estimated parameters $\mathbf{v}$ and $\bm \omega$ which are independent of the camera poses
increases the 
degree-of-freedom of the model compared to the interpolation methods. However, this makes the bundle adjustment prone to getting trapped in local minima due to the lack of additional constraints, especially when the prior camera pose is inaccurate. Inspired by the multi-stage methods \cite{dollar2010cascaded, jarrett2009best, zamir2021multi, fu2023cbarf} that use the iterative refinement approach to enhance the performance of optimization, we introduce a multi-stage strategy that performs coarse-to-fine optimization in series. As shown in Fig. \ref{fig:framework}, the camera poses estimated from the coarse stage are utilized as the initialization for the fine stage. In the coarse-stage, the input images are downsampled to increase the receptive field of the sampled points, thereby accelerating the convergence of the optimization. The features encoded through Tri-Mip encoding are {then} fed into an MLP network to obtain the color and volume density.  The initial inaccurate camera motion parameters and network parameters are optimized by supervising the rendering output and the corresponding downsampled RGB image. Finally, the erroneously estimated camera poses falling into local optima 
{are} identified and replaced to ensure the effectiveness of the coarse optimization 
(\cf Sec.~\ref{sec:Erroneous pose detection}). In the fine-stage, origin resolution images and the estimated camera poses from the coarse-stage are used for the final training. To learn the fine details as well as the rolling shutter effect in the image, we reinitialize the learning rates of parameters of MLP and mipmaps. The final optimized network model and camera parameters are obtained by supervising the rendering output and the original resolution RGB image.

\vspace{-1mm}
\subsection{Erroneous Pose Detection} \label{sec:Erroneous pose detection}

{
In the optimization of the coarse-stage, estimated camera poses that are grossly wrong lead to an erroneous optimization thus preventing the model from self-correction in the following fine-stage.}
{Unfortunately,} existing methods \cite{huynh2008scope, wang2004image, zhang2018unreasonable} that 
{assess} the quality of the pose estimation by inferior rendering quality 
{cannot be used} due to the rolling shutter effect in the image. Consequently, we introduce the rolling shutter epipolar geometry constraints \cite{dai2016rolling} as evaluation metrics to detect erroneously estimated poses. Given a pair of 
matched points $\mathbf{x}_i = \left[u_i, v_i, 1\right]$ and $\mathbf{x}_j = \left[u_j, v_j, 1\right]$, the rolling shutter epipolar error can be written as \cite{dai2016rolling}:
\begin{equation}
e_{ij} = \left[u_i, v_i, 1\right]\mathbf{K}^{-\top}\left[ \mathbf{t}_0^{ij} + u_i \mathbf{v}^i - u_j \mathbf{R}^{ij} \mathbf{v}^j
\right]_{\times} \mathbf{R}^{ij} \mathbf{K}^{-1}\left[u_j, v_j, 1\right]^{\top},
\end{equation}
where $\mathbf{R}^{ij} = \left(\mathbf{I} + [\bm{\omega}^i]_{\times} {u_i}\right) \mathbf{R}_0^{ij}  \left(\mathbf{I} -[\bm{\omega}^j]_{\times} {u_j}\right) $. $\mathbf{R}_0^{ij}$ and $\mathbf{t}_0^{ij}$ represent the relative rotation and translation of camera poses, respectively.  $\mathbf{K}$ is the calibrated camera intrinsic matrix and assumed to be known. To construct the epipolar constraint, we extract keypoints for each rolling shutter image using SuperPoint \cite{detone2018superpoint}, and obtain feature matches for each candidate image pair using SuperGlue \cite{sarlin2020superglue}. Only the matching point pairs from nearby views are utilized to reduce the number of the wrong matches. 
{We construct} a scene graph 
to obtain the nearby views. Two images become neighbors when they share enough image keypoint matches. We simply select nearby views by sorting their neighbors according to the number of matches in descending order. The matched points 
{are} identified as outliers when $e_{ij}$ exceeds $\delta_{th}$. 
{Moreover,} the camera poses are regarded as low-quality if the number of outliers exceeds a certain ratio and 
{are} replaced by the neighboring poses with {the} interpolation method \cite{li2023usb}. 
The scene graph construction only needs to be executed once for each scene 
{in} a preprocessing step.

\vspace{-1mm}
\section{Experiments}
\vspace{-1mm}
{We evaluate the effectiveness of our method on synthetic and real datasets.}
The performance of rolling shutter effect removal and novel view image synthesis {are benchmark} with the commonly used metrics: PSNR, SSIM and LPIPS between the recovered global shutter images and the ground truth global shutter images. In addition, we also conduct a quantitative assessment of newly generated images with various rolling shutter effects. For camera pose estimation, we perform a $\operatorname{Sim}(3)$ alignment against the ground truth trajectory to get the absolute trajectory error (ATE).  The Root Mean Square Error (RMSE) of the translation and rotation part is used for evaluation. 

\noindent\textbf{Baseline.} We compare our method against {the} learning-free method DiffSfM \cite{zhuang2017rolling} which constructs SfM similar to our rolling shutter modeling to remove the rolling shutter effect, and several learning-based methods: NeRF \cite{mildenhall2021nerf}, BARF \cite{lin2021barf}, Tri-MipRF \cite{hu2023Tri-MipRF} and USB-NeRF \cite{li2023usb}. For NeRF, BARF and Tri-MipRF, we assume the inputs are global shutter images to train the implicit radiance fields. 
{We re-implemented the interpolation method introduced in USB-NeRF based on Tri-MipRF to maintain consistency of the backbone. We denote our re-implementation of USB-NeRF as USB-NeRF-RE.}

\noindent\textbf{Datasets.}
{We use} the synthetic dataset
WHU-RS 
\cite{cao2020whu}, and real datasets ZJU-RS 
\cite{jinyu2019survey}
to verify the effectiveness of our method.  The WHU-RS dataset contains rolling shutter images and time-synchronized global shutter images and accurate ground truth collected in an ordinary room. To compare the impact of different rolling shutter effects, the datasets are divided into two trajectories with three sequences of different motion speeds (i.e. slow, medium, and fast corresponding to different rolling shutter effects). The scanline readout time is approximately 69.44$\mu$s. ZJU-RS datasets are collected with two mobile phones.  Since the datasets lacks corresponding GS images, we choose two sequences and only evaluate the accuracy of the recovered camera trajectories compared with the groundtruth trajectories. The scanline readout time is approximately 20.83$\mu$s. We also conducted a qualitative analysis of rendering quality in a LivingRoom scene \cite{li2023usb} and new  RS dataset generation by setting different camera motions on the LLFF dataset \cite{mildenhall2019local}.

\noindent\textbf{Experimental Settings.}
We parameterize the camera poses $\mathbf{T}$ with $\mathfrak{se}(3)$  and initialize all the camera poses with the ground truth. To simulate inaccurate camera poses, the Gaussian noises with standard deviation $\delta_{trans} = 0.10$m and $\delta_{rot} = 1.15^\circ$  are added to the translation part and rotation part of the initial poses, which is similar to  \cite{lin2021barf}.  As {the} whole sequence of {the} WHU-RS dataset, ZJU-RS dataset are too long for NeRF to process, we choose a subset {of} frames for each sequence. The initial {linear velocity} $\mathbf{v}$ and angular velocity $\bm \omega$ of each pose are set to zero.  The camera intrinsic and scanline readout time of the RS camera are assumed known (provided by the dataset). To test the performance of different methods on unordered frames, we also randomly shuffled the selected sequences to generate the unordered datasets. 

\subsection{Ablation Experiments}
\begin{figure}[b!]
  \centering
  \includegraphics[width=0.95\textwidth]{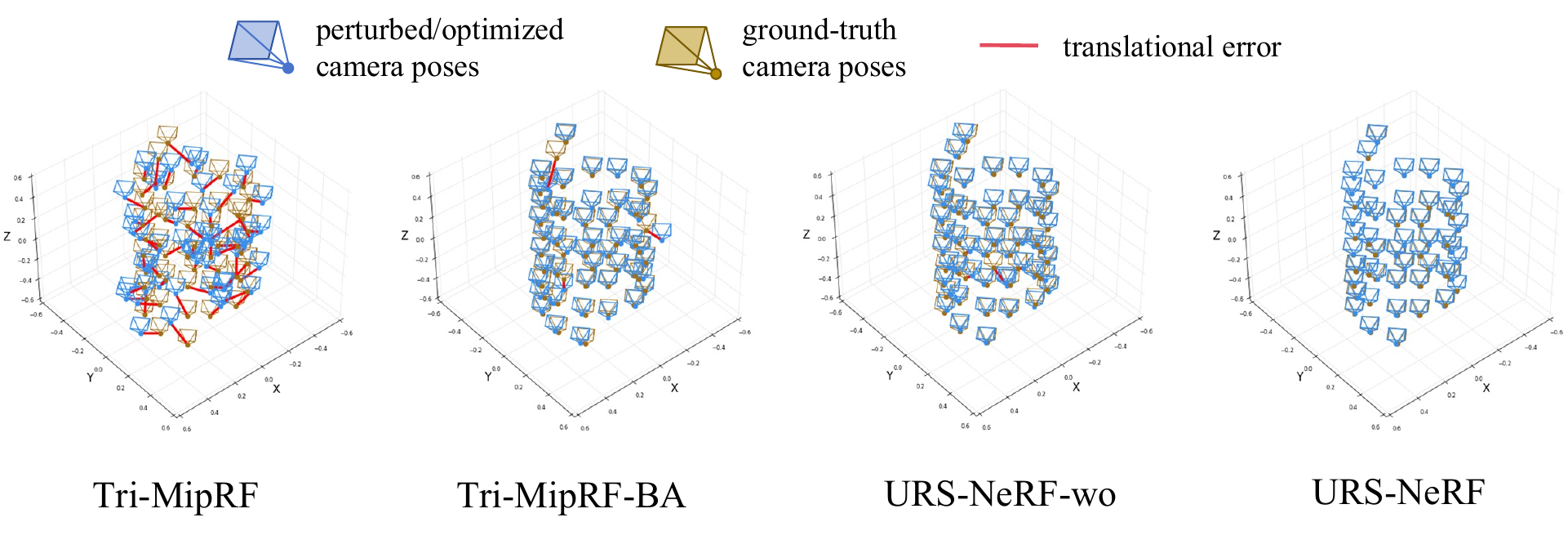}
  \caption{Visual comparison of the initial and optimized camera poses for Tri-MipRF, Tri-MipRF-BA, URS-NeRF-wo, and URS-NeRF on Traj1-medium-large setting. URS-NeRF successfully realigns all the camera frames, while some estimated poses of Tri-MipRF-BA and URS-NeRF-wo get stuck at suboptimal solutions. The original Trimip-RF lacks the bundle adjustment, resulting in the poorest performance.}
  \label{fig:ablation_pose_estimation}
\end{figure}

To illustrate the effect of the introduced bundle adjustment in 
Tri-MipRF, the superiority of our rolling shutter modeling method, as well as the behavior of the coarse-to-fine strategy, we conducted ablation experiments to evaluate the performance of rolling shutter removal for the training view. Methods used for comparison include: Tri-MipRF is the baseline, Tri-MipRF-BA introduces the bundle adjustment in the Tri-MipRF, URS-NeRF-wo removes the coarse-to-fine strategy used in the  URS-NeRF and URS-NeRF. As shown in Tab.\ref{tab: ablation_experiments},  URS-NeRF exhibits superior performance on both small perturbation settings and large perturbation settings, which means our method formulates the physical image formation process of the RS camera, meanwhile, it also verifies the effectiveness of our coarse-to-fine strategy. Specially, by comparing the experimental results of Tri-MipRF and Tri-MipRF-BA, we can find that introducing the bundle adjustment into Tri-MipRF can significantly improve the quality of 3D scene reconstruction and the accuracy of the pose estimation, which also can be shown in Fig. \ref{fig:ablation_pose_estimation}. Then from the experiment results of Tri-MipRF-BA and USB-NeRF-wo, we can see that after modeling the RS effect in the image, the quality of rendering is further improved. Finally, by comparison of URS-NeRF-wo and URS-NeRF, we can observe that URS-NeRF employs the coarse-to-fine approach, enabling the detection and correction of poses trapped in local minima,  which leads to more precise poses and enhances the rendering quality. In the large and small settings, compared to URS-NeRF-wo, the PSNR improves by 8.13$\%$ and 1.13$\%$, respectively.

\begin{table}[t!]
\captionsetup {font={small,stretch=0.5}}
\caption{Ablation experiments for Tri-MipRF, Tri-MipRF-BA, URS-NeRF-wo and URS-NeRF on Traj1-medium scene of WHU-RS dataset. Traj1-medium-large and Traj1-medium-slow are subjected to Gaussian noise with standard deviation $\delta_{trans} = 0.30$ m,  $\delta_{rot} = 1.15^\circ$  and $\delta_{trans} = 0.10$ m
$\delta_{rot} = 1.15^\circ$ rad, respectively.  
For each metric, the best in \textbf{bold}.} 
\vspace{-2.5em}
	\label{tab: ablation_experiments}
	\begin{center}
	\setlength\tabcolsep{9pt}
	\setlength{\belowcaptionskip}{0pt}
  \renewcommand\arraystretch{1.2}
	\scriptsize
	\resizebox{\linewidth}{!}{
		\begin{tabular}{c|ccc|ccc}
			\toprule
			&  \multicolumn{3}{|c}{Traj1-medium-large}  &  \multicolumn{3}{|c}{Traj1-medium-small} \\
			& PSNR$\uparrow$ & SSIM$\uparrow$ & LPIPS$\downarrow$ & PSNR$\uparrow$ & SSIM$\uparrow$ & LPIPS$\downarrow$ \\
			\midrule
			Tri-MipRF      &15.24 &0.46 &0.70 &16.04 &0.45 &0.64\\
			Tri-MipRF-BA      &26.35 &0.84 &0.13 &27.13 &0.86 &0.11\\
    		URS-NeRF-wo    &27.16 &0.84 &0.11 &29.01 &0.90 &0.07\\
			URS-NeRF        &\bf 29.37 & \bf0.90 & \bf0.05 &\bf 29.34 & \bf0.91 &\bf 0.06\\
			\specialrule{0.08em}{1pt}{1pt}
		\end{tabular}
		}
	\end{center}
        \vspace{-3em}
\end{table}

\subsection{Quantitative Experiments}
We evaluate the performance of our method against state-of-the-art methods in terms of rolling shutter removal and the accuracy of trajectory estimation. Tab. \ref{tab: rs removal training view}  presents comparisons between the groundtruth global shutter images and the synthetic images without RS effect from the training view.  URS-NeRF achieves the best performance on the fast, medium, and slow modes of the unordered images.  This is because our method effectively models the RS effect in the image, resulting in the implicit 3D scene representation that removes the RS effect. Since NeRF and Tri-MipRF do not incorporate bundle adjustment, the inaccurate poses and rolling shutter effects in the images result in poor rendering quality. Furthermore, comparing NeRF with BARF, and Tri-MipRF with Tri-MipRF-BA, we can observe that the rendering quality of Tri-MipRF-BA is improved compared to Tri-MipRF. However, the rendering quality of BARF is worse than that of NeRF. This is because the poses estimated by BARF fall into a local minimum when training the images for trajectory 2.
\begin{table}[t!]
    \scriptsize
	\captionsetup {font={small,stretch=0.5}}
    \caption{Quantitative comparisons on the synthetic datasets in terms of rolling shutter effect removal for training view on WHU-RS dataset. For the fast, medium, and slow modes of the WHU-RS dataset, the average values of each metric are computed from two scenes.  For each metric, the best in {\textbf{bold}} for the unordered datasets and \textcolor{blue}{blue} for the sequence video datasets.}
    \vspace{-2mm}
	\label{tab: rs removal training view}
	\begin{center}
	\setlength\tabcolsep{2pt}
	\setlength{\belowcaptionskip}{0pt}
    \renewcommand\arraystretch{1.2}
	\scriptsize
 	\vspace{-1.0em}
	\resizebox{\linewidth}{!}{
		\begin{tabular}{c|c|ccc|ccc|ccc}
			\toprule
			&&  \multicolumn{3}{|c}{WHU-RS-Fast}  &  \multicolumn{3}{|c}{WHU-RS-Medium}  &  \multicolumn{3}{|c}{WHU-RS-Slow} \\
			& & PSNR$\uparrow$ & SSIM$\uparrow$ & LPIPS$\downarrow$ & PSNR$\uparrow$ & SSIM$\uparrow$ & LPIPS$\downarrow$ & PSNR$\uparrow$ & SSIM$\uparrow$ & LPIPS$\downarrow$\\
			\midrule
			\multirow{6}{*}[0pt]{Un-view}& NeRF      & 21.66& 0.57 & 0.67 & 20.89 & 0.56 & 0.69 & 21.05 & 0.56 & 0.67\\
			&BARF  & 16.27 & 0.49 & 0.64 & 17.65& 0.53 & 0.58 & 15.70 & 0.47 & 0.63\\
		      &DiffSfM   &24.86 & 0.81 & 0.18& 26.75 & 0.86 & 0.12 & 27.71& 0.87 & 0.10\\
			&Tri-MipRF &16.34 & 0.47& 0.61 & 16.80 & 0.50& 0.57&16.91 & 0.49 & 0.57 \\
            & Tri-MipRF-BA &19.80 & 0.60& 0.57 & 19.08 & 0.58& 0.62&20.33 & 0.62 & 0.60 \\
            &USB-NeRF-RE  & 16.91 & 0.50 & 0.61 &  19.18 &  0.60 &  0.50 &  21.07 &0.65 & 0.42\\
			&URS-NeRF & \textbf{27.55} &\textbf{0.85} & \textbf{0.10} &{\textbf{28.90}} &{\textbf{0.88}} &{\textbf{0.08}} & {\textbf{29.42}} &{\textbf{0.89}} &{\textbf{0.08}}\\
			\specialrule{0.08em}{1pt}{1pt}
            \multirow{2}{*}{Seq-view}&USB-NeRF-RE  &\textcolor{blue}{29.07} & \textcolor{blue}{0.87} & \textcolor{blue}{0.12} & \textcolor{blue}{29.54} & \textcolor{blue}{0.88} & \textcolor{blue}{0.11} & \textcolor{blue}{30.16}&\textcolor{blue}{0.89} &\textcolor{blue}{0.10} \\
			&URS-NeRF & 27.70 &0.85 & 0.14&28.97&0.87 &0.11 & 29.41 & 0.88 & 0.11\\
            \specialrule{0.08em}{1pt}{1pt}
		\end{tabular}
	
		}
	    \vspace{-3em} 
	\end{center}
\end{table}
\begin{table}[b!]
    \scriptsize
        \vspace{-1.5em}
	\captionsetup {font={small,stretch=0.5}}
	\caption{Camera pose estimation on unordered view dataset and sequence view dataset. We evaluate the average values of the translation error (m) and rotation error ($^\circ$). For each metric, the best in \textbf{bold} for unordered datasets and \textcolor{blue}{blue} for the sequence video datasets.}
    \vspace{-2mm}
	\label{tab: pose estimation accuracy main}
	\begin{center}
	\setlength\tabcolsep{4pt}
    \renewcommand\arraystretch{1.2}
	\setlength{\belowcaptionskip}{0pt}
	\scriptsize
 	\vspace{-2.0em}
	\resizebox{\linewidth}{!}{
		\begin{tabular}{c|c|cc|cc|cc|cc}
			\toprule
			& &  \multicolumn{2}{|c}{WHU-RS-Fast}&  \multicolumn{2}{|c}{WHU-RS-Medium}&  \multicolumn{2}{|c}{WHU-RS-Slow}&  \multicolumn{2}{|c}{ZJU-RS}\\
		    & & Trans & Rot & Trans & Rot &  Trans & Rot & Trans & Rot  \\
			\midrule
		    \multirow{6}{*}[0pt]{Un-view}&  BARF      & 0.041& 3.534 & 0.150 & 8.373 & 0.092 & 1.967&  0.055 & 5.653\\
		      & DiffSfM   &0.020 & 0.962 &  0.011& 0.839& \bf0.009& \bf 0.386 &0.013 &\bf 1.237\\
		    & Tri-MipRF-BA &0.075 & 4.846& 0.127 & 4.088 & 0.108& 3.138&0.011 &2.009  \\
            & USB-NeRF-RE  &0.495& 10.787 & 0.375 & 7.743 & 0.304 & 8.511  &{0.124} &{15.912}\\
		      & URS-NeRF & \bf0.014 & \bf0.697& \bf0.009 &\bf 0.381& 0.013 & 0.472  &\bf{0.007} &{2.871}\\
			\specialrule{0.08em}{1pt}{1pt}
            \multirow{2}{*}{Seq-view} &  USB-NeRF-RE  &{\textcolor{blue}{0.008}} &0.649 &0.009 & 0.324 & 0.011 & {\textcolor{blue}{0.296}}   &0.009 &2.462 \\
		    & URS-NeRF &{0.016} &{\textcolor{blue}{0.600}} &{\textcolor{blue}{0.006}} &{\textcolor{blue}{0.310}}  &{\textcolor{blue}{0.009}} &{0.326} &{\textcolor{blue}{ 0.007}} &{\textcolor{blue}{1.732}}\\

        \specialrule{0.08em}{1pt}{1pt}
		\end{tabular}
		}
	\vspace{-2em}
	\end{center}
\end{table}
Meanwhile, our method also outperforms DiffSfM, which employs the same rolling shutter modeling method as ours. However, DiffSfM only utilizes two frames to remove the RS effect, while our method uses images from multi-views, resulting in better performance. Finally, by comparing the results of USB-NeRF-RE and URS-NeRF, our method works on both unordered and sequential data, while USB-NeRF-RE fails to reconstruct the 3D scene on unordered data. Meanwhile, the difference in RS effect removal between our method and USB-NeRF on sequential data is marginal. This also demonstrates the generalization of our method.  It's worth noting that our method performs best in the low-speed mode. As the camera movement speed increases, the rendering accuracy deteriorates. This is because slower speeds align more accurately with the assumption of constant velocity, resulting in better modeling performance.

Table \ref{tab: pose estimation accuracy main} presents the camera motion trajectory estimation results with both synthetic and real datasets. The results demonstrate that both BARF and Tri-MipRF-BA suffer from the rolling shutter effect. The introduced distortions would affect the estimation of the poses and may even make the estimated pose worse. On the contrary, since DiffSfM and URS-NeRF formulate the physical image formation process of RS camera when training the implicit neural radiance field, the accuracy of the pose estimation is improved. From the pose estimation results, it is also evident that the cubic interpolation method used by USB-NeRF-RE imposes strict limitations on the input data, hence being unable to handle unordered images. However, our method is not affected by the data order, making it more flexible for practical applications.

\begin{figure}[t!]
  \centering
  \includegraphics[width=\textwidth]{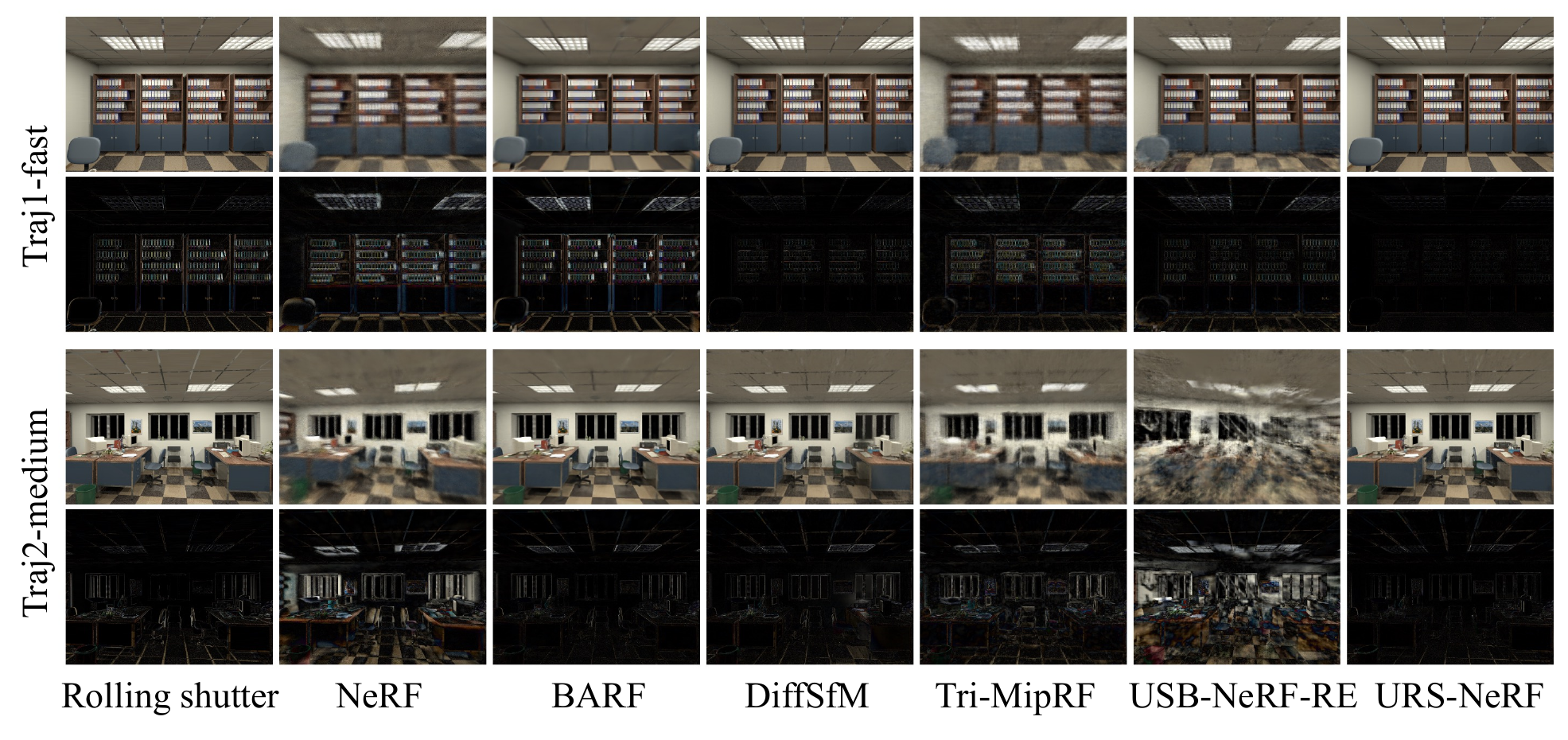}
  \vspace{-5mm}
  \caption{ Qualitative comparisons with the unordered images on the WHU-RS datasets. The second row consists of disparity maps between rendered images and ground truth, where darker areas indicate better performance. The experiments demonstrate a significant improvement in both rendering quality and reduction of rolling shutter effects with our method.}
  \label{fig:main_experiment_render}
  \vspace{-4mm}
\end{figure}

\begin{figure}[t!]
  \centering
  \includegraphics[width=\textwidth]{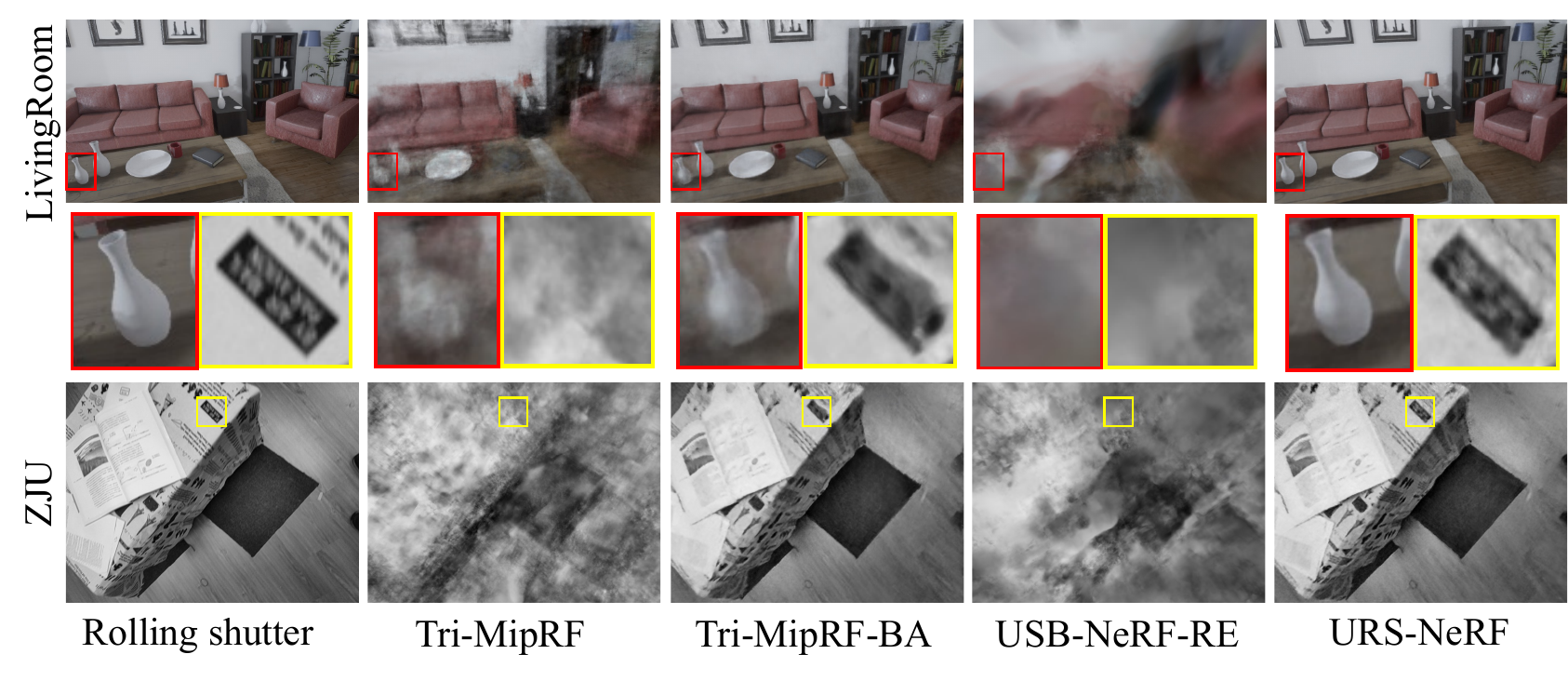}
    \vspace{-5mm}
    \caption{ Qualitative comparisons on ZJU and LivingRoom datasets. The detailed and overall images demonstrate that our method achieves better performance compared to other works on unordered images. }
  \label{fig:vis_ZJU}
  \vspace{-5mm}
\end{figure}

\begin{figure}[h]
  \centering
  \vspace{-1mm}
  \includegraphics[width=0.9\textwidth]{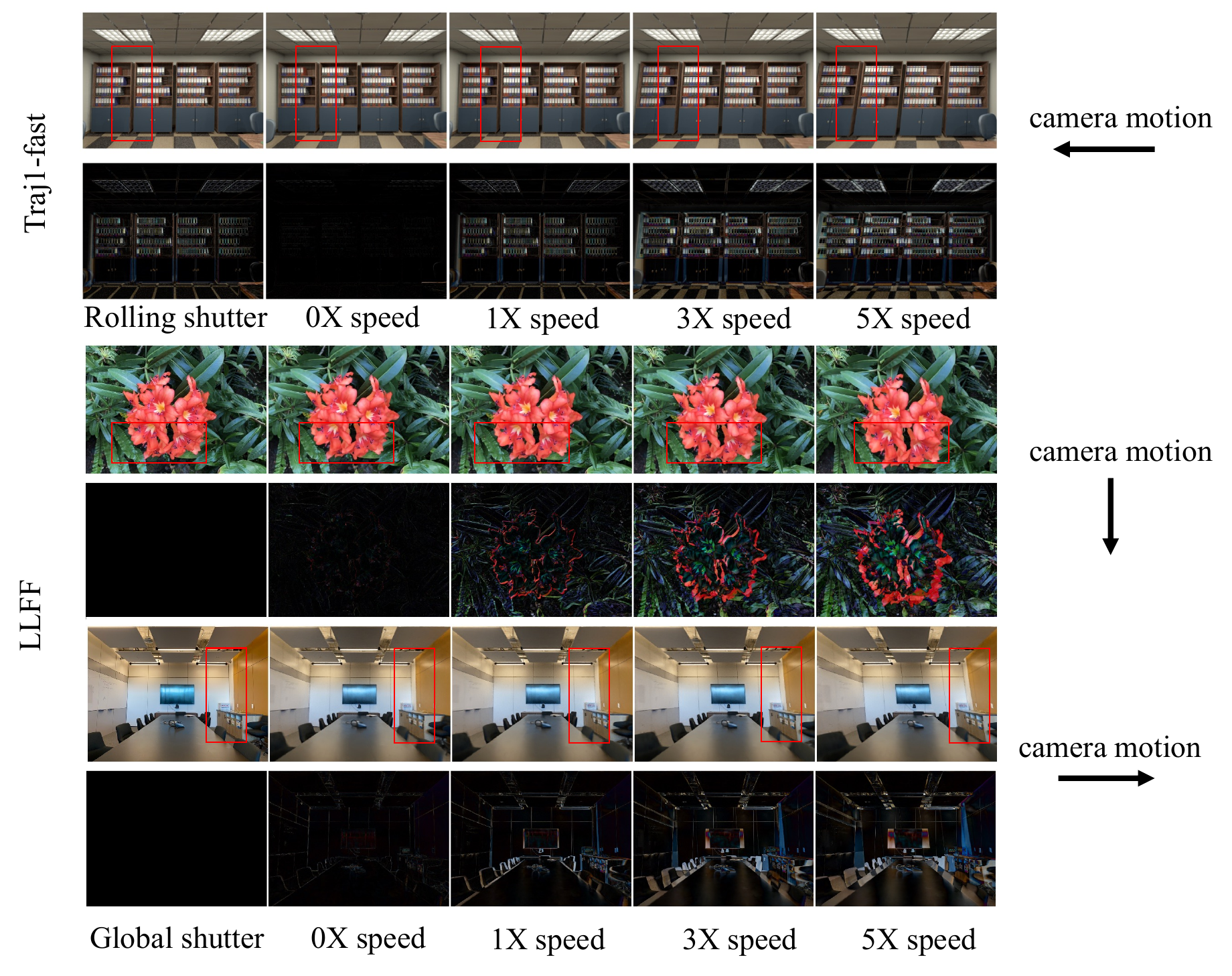}
   \vspace{-1em}
   \caption{New rolling shutter datasets generation with WHU-RS datasets. Our method can not only remove the rolling shutter effect ($2^{\rm th}$ cols) from the rolling shutter images but also synthesize datasets with varying degrees of rolling shutter effects (the camera motion speeds are 1x, 3x, and 5x, respectively). }
  \label{fig:new_data_generation}
  \vspace{-5mm}
\end{figure}

\subsection{Qualitative Experiments}
We also evaluate the qualitative performance of our method against the other baseline methods. Fig. \ref{fig:main_experiment_render} presents the comparisons with WHU-RS dataset.  From both trajectory 1 and trajectory 2 scenarios, it can be seen that our method can obtain a 3D representation that is not affected by the rolling shutter effect under both fast and medium-speed camera movements, thereby synthesizing global shutter images. NeRF, BARF, and Tri-MipRF fail to learn the underlying undisturbed 3D scene representation, which proves the necessity to properly model the physical image formation process of RS camera when training with rolling shutter images.  By comparing the rendering results of USB-NERF-RE and IRS-NERF, we can see that USB-NeRF is unable to handle unordered input data and the rendering quality deteriorates as the camera movement speed increases. 
More rendering results can be found in Fig. \ref{fig:vis_ZJU}.

Since our method models the physical image formation process of a RS camera by estimating the camera's velocity and angular velocity, we are not only able to train a 3D scene representation unaffected by RS effect but also synthesize new images with various levels of RS effects. As shown in Fig. \ref{fig:new_data_generation}, we not only restore the GS images but also generate images with different degrees of RS effects using the estimated velocity and angular velocity. Furthermore, we can also train an implicit neural field using GS images and synthesize images with varying RS effects by setting different camera movement speeds.
More novel view image synthesis and RS dataset generation results can also be found in the Appendix and supplementary video. They also demonstrate the superior performance of our method over prior works.

\vspace{-2mm}
\section{Conclusion}
\vspace{-2mm}
In this paper, we presented unordered rolling shutter bundle adjustment for neural radiance fields. The method introduces the bundle adjustment into Tri-MipRF to estimate the RS camera pose,  velocity and implicit 3D representation. To prevent the bundle adjustment with the rolling shutter model from getting stuck in local minima, we adopt a coarse-to-fine strategy and erroneous pose detection.  Experimental results demonstrate that our method can successfully learn the true underlying 3D representations and recover the motion trajectory, given unordered input RS images. Our method offers greater flexibility compared to interpolation-based approaches, providing a novel solution for implicit reconstruction with RS images.

\clearpage  
\title{ Supplementary } 

\author{ }

\institute{ }

\maketitle
\appendix
\renewcommand\thefigure{S\arabic{figure}}
\renewcommand\thetable{S\arabic{table}}

\section{Implementation and Training Details}

We implemented our method in PyTorch, running a total of 25K steps on a computer with  Intel i7-9750H$@$2.6GHz CPU and NVIDIA RTX 4090 GPU. The coarse stage comprised 10K iterations, followed by the fine stage with 15K iterations. The Adam optimizer \cite{diederik2014adam} is used to estimate the weights of the network, embedding pose parameters, and velocity parameters. For Tri-Mip $\mathcal{M}$, the learning rate was set to $2 \times 10^{-3}$, while for MLP, it was $2 \times 10^{-2}$. And the learning rates for camera pose and velocity were set to $2 \times 10^{-3}$ and $2 \times 10^{-4}$, respectively. We followed a learning rate reduction schedule, decreasing it by $0.6\times$ at $\frac{1}{2}$, $\frac{3}{4}$, $\frac{5}{6}$, and $\frac{9}{10}$ of the total steps, consistent with \cite{hu2023Tri-MipRF}. 
The total training iterations for NeRF \cite{mildenhall2021nerf} and BARF \cite{lin2021barf} are 200K. The pose corresponding to the first row of each image is assumed as the pose of the frame. And the pose accuracy is evaluated by the tool evo \cite{grupp2017evo}. Since  DiffSfM \cite{zhuang2017rolling} cannot synthesize novel images, we first apply the method to restore global shutter images and then use them as input to train Tri-Mip-BA.

\section{Details on Selected datasets}
We use the synthetic datasets WHU-RS \cite{cao2020whu}, and real datasets ZJU-RS \cite{jinyu2019survey} to verify the effectiveness of our method. We conducted experiments using 6 sequences from the WHU-RS dataset, comprising two scenarios, with each scenario including fast, medium, and slow sequences. To analyze the performance of our method under different camera motion speeds, we selected similar scenes and trained the model using approximately 100 images for each sequence.  For the ZJU-RS dataset, we selected 6 sequences (D0, D2, D3, D8, C5, C11) out of the 23 available for reconstruction, all of which were captured using smartphones equipped with rolling shutter cameras. For each sequence, we selected 70-100 images for reconstruction. In the main paper, we analyzed the performance of different methods on D0 and D2 sequences. In the supplementary materials, more experimental results are reported.

\section{Supplementary  Analysis}
\subsection{Training Time Analysis}
As described in the main paper, since USB-NeRF \cite{li2023usb} is based on BARF \cite{lin2021barf}, we re-implemented the interpolation method used in USB-NeRF based on Tri-MipRF \cite{hu2023Tri-MipRF} to maintain consistency of the backbone. We tested the training time and querying time of the methods used in the main paper, especially comparing the time consumption between the interpolation method and our method with the same backbone. From Tab. \ref{tab: time consumption}, we can observe that the training and querying progress of  NeRF and BARF is particularly slow due to the adoption of the coordinate-based MLPs in the network. Thanks to the Tri-Mip representation in Tri-MipRF \cite{hu2023Tri-MipRF}, Tri-MipRF and its extensions can achieve both high-fidelity anti-aliased renderings and efficient reconstruction. The training and querying speeds have been significantly improved. Specifically, by comparing  Tri-MipRF and Tri-MipRF-BA, it can be observed the training time doubled nearly after introducing bundle adjustment. Then, the computation time of URS-NeRF is longer than Tri-MipRF-BA when estimating additional velocities.  Finally, the training and querying time of USB-NeRF-RE is the longest, mainly due to the complex cubic interpolation calculation and the differentiation in the backpropagation process. It is worth noting that the training time of the DiffSfM is similar to Tri-MipRF-BA. However, DiffSfM requires an additional 22859.34s to eliminate the rolling shutter effect in the images.

\begin{table}[t!]
	\captionsetup {font={small,stretch=0.5}}
	\caption{Average training and querying time consumption of Traj1-fast scene of WHU-RS dataset in seconds.} 
	\label{tab: time consumption}
	\begin{center}
	\setlength\tabcolsep{2pt}
	\setlength{\belowcaptionskip}{0pt}
    \renewcommand\arraystretch{1.5}
	\scriptsize
 	\vspace{-2.0em}
	\resizebox{\linewidth}{!}{
		\begin{tabular}{c|ccccccc}
			\toprule
			 & NeRF&BARF& DiffSfM&Tri-MipRF& Tri-MipRF-BA & USB-NeRF-RE& URS-NeRF\\
			\midrule
	Training Time   & 32314.32 & 43174.51 &1014.29 (+22859.34)  & 543.29 &1010.88   &2092.35& 1407.25 \\
	Querying Time   & 9.25  & 9.31  &1.04 & 0.99 & 0.98 &1.30 & 0.98\\
             \specialrule{0.08em}{1pt}{1pt}
		\end{tabular}
		\vspace{-2em}
		}
	\end{center}
\end{table}

\begin{figure}[b!]
  \centering
  \includegraphics[width=\textwidth]{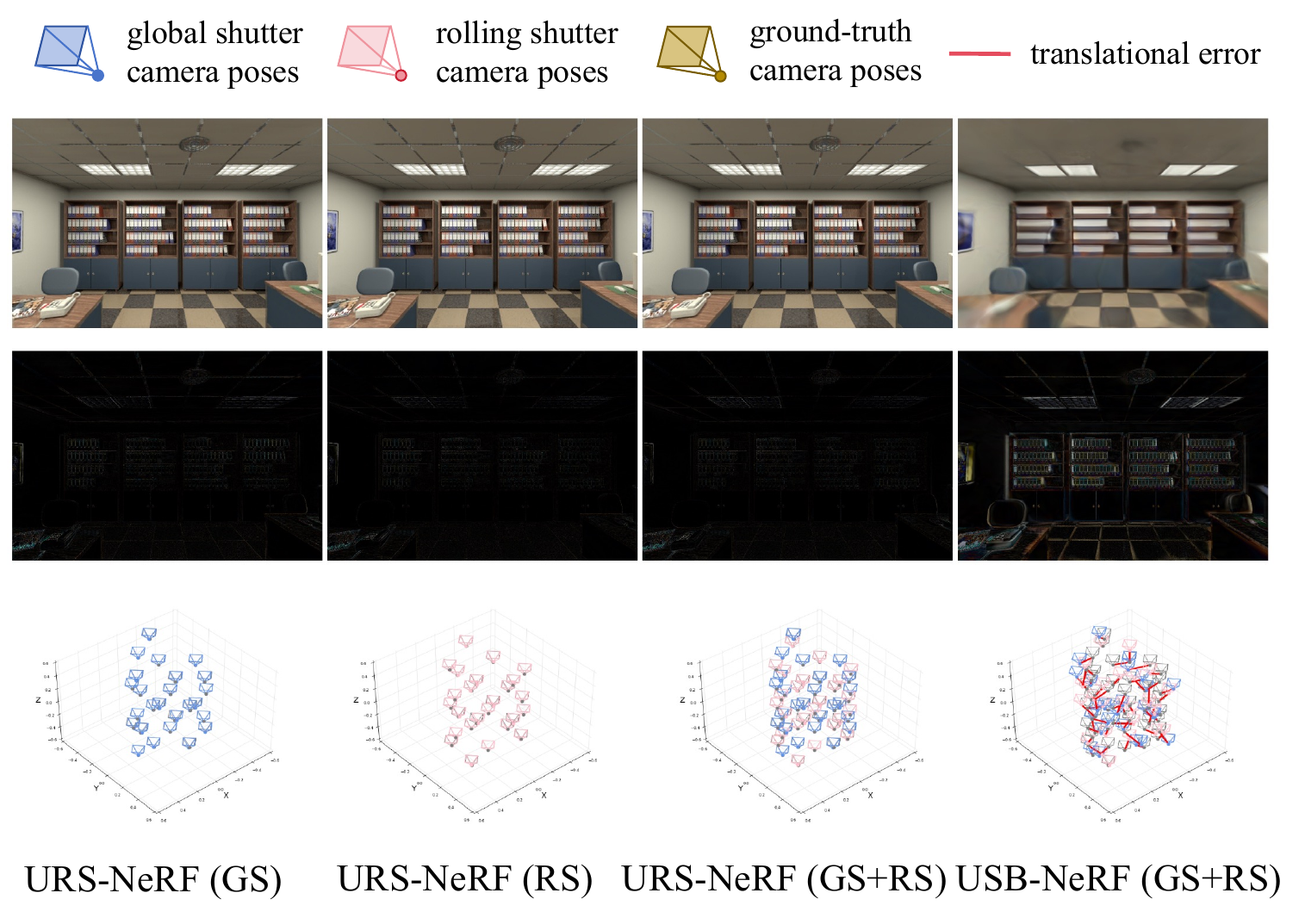}
  \caption{Given unordered GS/RS images, our model can simultaneously learn the undisturbed 3D scene representation and recover the unordered camera poses with only GS images ($1^{\rm th}$ col), only RS images ($2^{\rm th}$ col) and mixed RS, GS images ($3^{\rm th}$ col). However, USB-NeRF can only take sequential images as input and cannot process hybrid images of RS and GS simultaneously, which limits its practical applications (e.g. reconstruction using different types of cameras or utilizing crowdsourced data for reconstruction).  The second row presents residual images (the darker the better) that are defined as the absolute difference between the rendered undisturbed images (first row) and ground truth global shutter images.}
  \label{fig:abstract_figure}

\end{figure}

\begin{table}[b!]
	\captionsetup {font={small,stretch=0.5}}
	\caption{Quantitative comparisons on the synthetic datasets regarding novel view synthesis on WHU-RS dataset. For the fast, medium, and slow modes of the WHU-RS dataset, the average values of each metric are computed from two scenes. For each metric, the best in {\textbf{bold}} for the unordered datasets and \textcolor{blue}{blue} for the sequence video datasets.} 
	\label{tab: rs removal novel view}
	\begin{center}
	\setlength\tabcolsep{2pt}
	\setlength{\belowcaptionskip}{0pt}
    \renewcommand\arraystretch{1.2}
	\scriptsize
 	\vspace{-2.0em}
	\resizebox{\linewidth}{!}{
		\begin{tabular}{c|c|ccc|ccc|ccc}
			\toprule
			& &  \multicolumn{3}{|c}{WHU-RS-Fast}  &  \multicolumn{3}{|c}{WHU-RS-Medium}  &  \multicolumn{3}{|c}{WHU-RS-Slow} \\
			& & PSNR$\uparrow$ & SSIM$\uparrow$ & LPIPS$\downarrow$ & PSNR$\uparrow$ & SSIM$\uparrow$ & LPIPS$\downarrow$ & PSNR$\uparrow$ & SSIM$\uparrow$ & LPIPS$\downarrow$\\
			\midrule
			\multirow{6}{*}[0pt]{Un-view}&NeRF      & 18.13 & 0.46& 0.72& 18.57 & 0.48 & 0.72 & 18.54 & 0.48 & 0.71\\
			& BARF      & 16.37 & 0.49 & 0.64 & 17.54 & 0.49 & 0.59& 15.34 & 0.46 & 0.64\\
		      & DiffSfM   & 25.07& 0.80& 0.19 & 27.08 & 0.85 & 0.12 & 27.78& 0.87& 0.11\\
			& Tri-MipRF &16.35 & 0.47& 0.61 & 16.80 & 0.49& 0.57&16.94 & 0.50 & 0.58\\
            & Tri-MipRF-BA &19.17 & 0.58 & 0.58 & 18.87 & 0.58& 0.63&19.77 & 0.60 & 0.61\\
            & USB-NeRF-RE  &16.64 &  0.49 &  0.61 &  18.76 & 0.58 &0.51 &  20.56&  0.64 & 0.43\\
			& URS-NeRF &\textbf{27.27} &\textbf{ 0.84} &\textbf{0.11} &\textbf{ 28.48} &\textbf{0.87} &\textbf{0.09} &\textbf{ 29.02} &\textbf{0.88} &\textbf{0.09}\\
			\specialrule{0.08em}{1pt}{1pt}
            \multirow{2}{*}{Seq-view} & USB-NeRF-RE  &\textcolor{blue}{28.93} &\textcolor{blue}{0.86} & \textcolor{blue}{0.13} & \textcolor{blue}{29.61} & \textcolor{blue}{ 0.88} & \textcolor{blue}{ 0.10} & \textcolor{blue}{29.85}&  \textcolor{blue}{0.89} & \textcolor{blue}{ 0.10}\\
			& URS-NeRF &{27.56} &{ 0.85} &{0.15} &{28.82} &{0.87} &{0.11} &{29.21} &{ 0.87} &{ 0.11}\\
             \specialrule{0.08em}{1pt}{1pt}
		\end{tabular}
		\vspace{1em}
		}
	
	\end{center}
\end{table}

\subsection{Generality Analysis}
We conducted further analysis of the generality of our method. As shown in Fig. \ref{fig:abstract_figure}, due to introducing the estimated parameters $\mathbf{v}$ and $\bm \omega$ which are independent of the camera poses,  our method can reconstruct the scene with only GS images, only RS images, and mixed RS, GS images. This indicates that our method is not only unaffected by the order of input images but also does not require restricting the types of images, which conforms to the generality of utilizing multi-source data for reconstruction in SfM.

\subsection{Additional Experimental Results}
We note that we have reported the quantitative comparisons on the training view on WHU-RS dataset in Tab.2, which indicates the effectiveness of removing the rolling shutter effect on the training view. In tab. \ref{tab: rs removal novel view}, we also evaluate the performance of our method against the state-of-the-art methods in terms of novel view synthesis. Some conclusions consistent with the main paper can be obtained. URS-NeRF still outperforms other methods on the unordered datasets. However, due to the sequential constraints used in USB-NeRF-RE,  USB-NeRF-RE can not handle unordered input images.

\begin{figure}[t!]
  \centering
  \includegraphics[width=\textwidth]{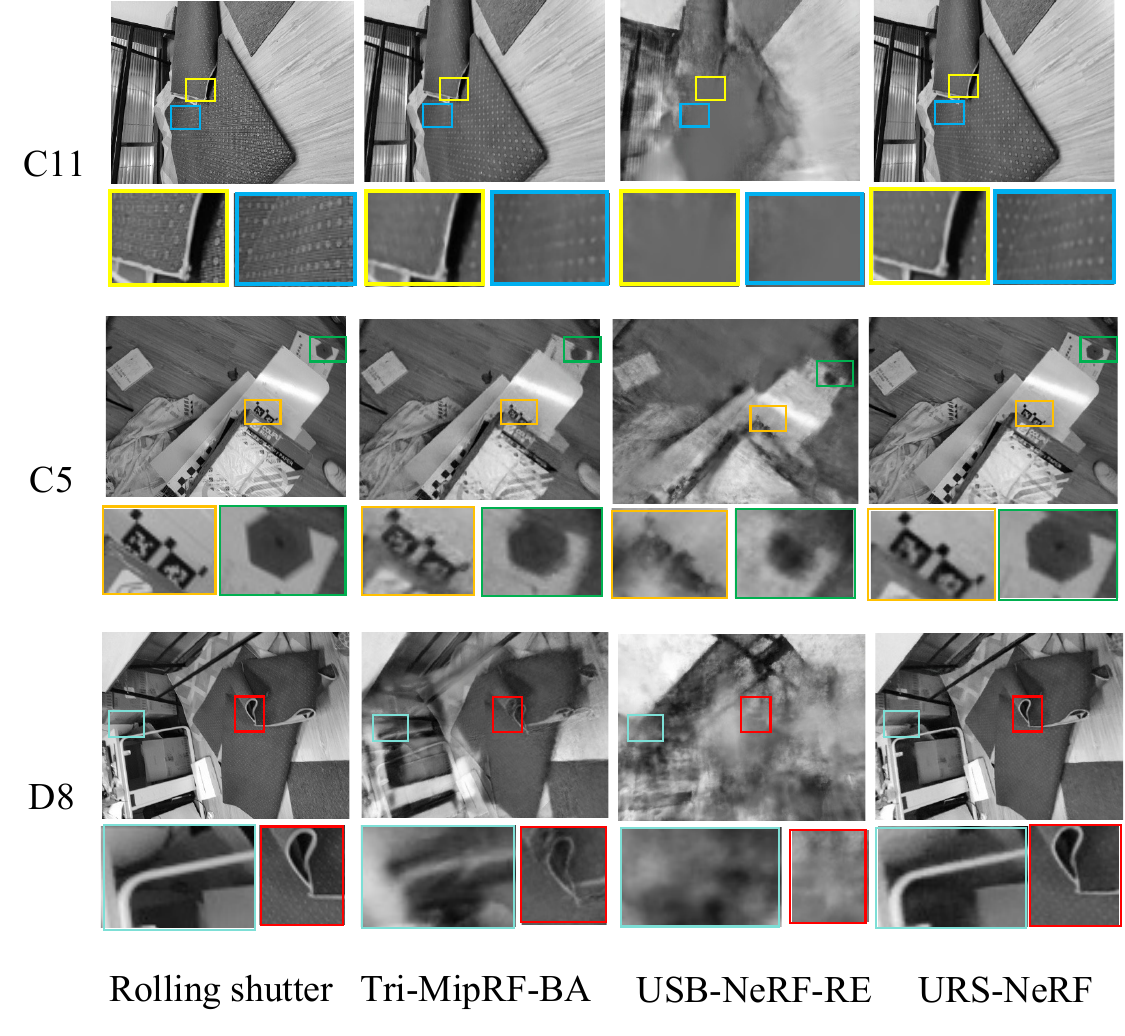}
  \caption{Qualitative comparisons on ZJU-RS datasets. The detailed and
overall images demonstrate that our method achieves better performance compared to
other works on unordered images.}
  \label{fig:zju_rendering}
  \vspace{-2.5em}
\end{figure}

\begin{figure}[t!]
  \centering
  \includegraphics[width=\textwidth]{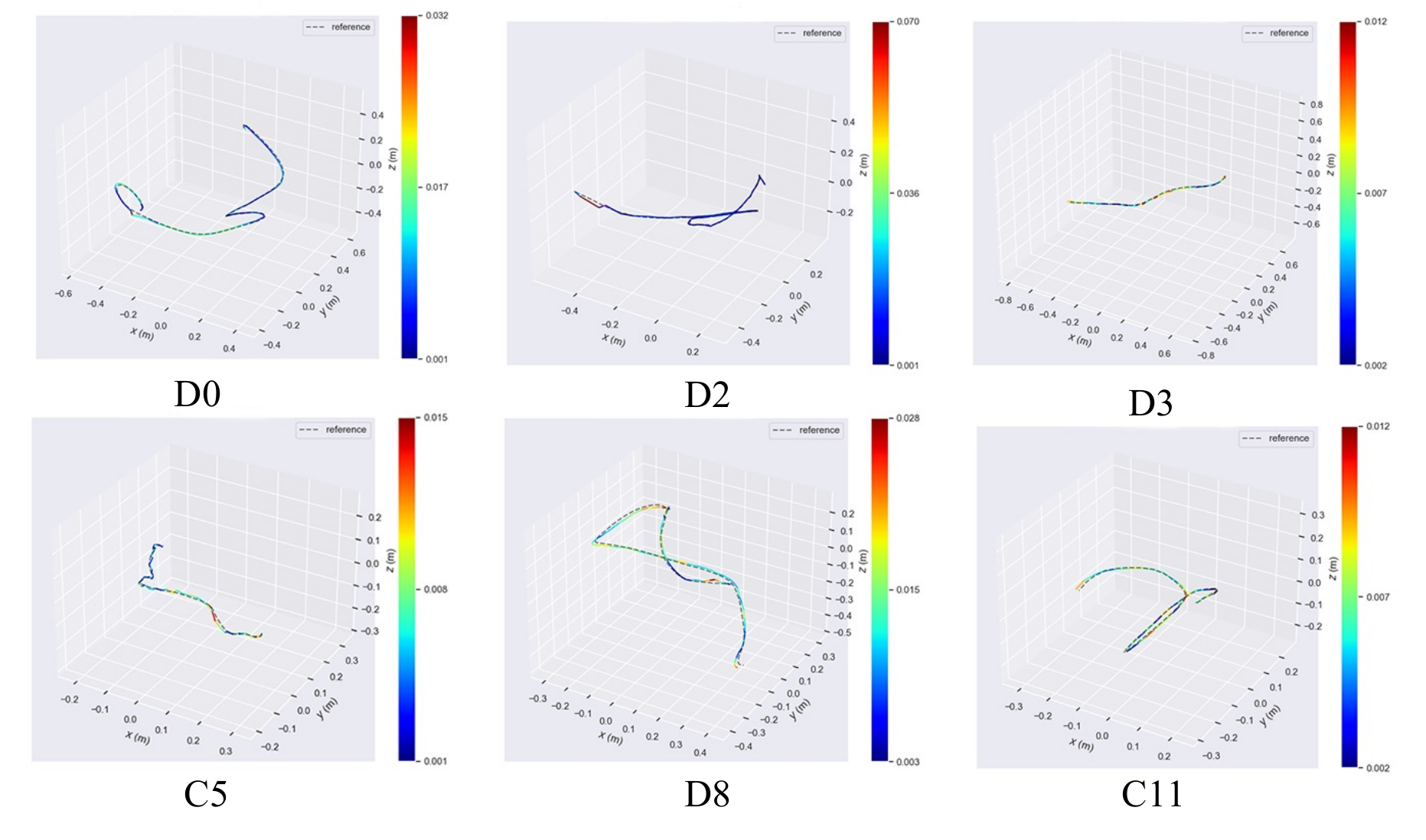}
  \caption{Comparisons of estimated trajectories of real ZJU-RS datasets. The experimental results demonstrate that our method can recover the motion trajectories of unordered images. }
  \label{fig:zju_traj}
\end{figure}

We also conducted additional quantitative and qualitative experimental analyses on the unordered view of real ZJU-RS datasets. Tab. \ref{tab: pose estimation accuracy supplementary} presents the accuracy of trajectory estimation using different methods. Fig. \ref{fig:zju_rendering} depicts the quality of rendering, while Fig. \ref{fig:zju_traj}  shows the recovered trajectories compared with the groundtruth.  These results demonstrate that our method effectively improves the performance of the reconstruction with the images captured by the smartphones.

\begin{table}[b!]
    \scriptsize
        \vspace{-1.5em}
	\captionsetup {font={small,stretch=0.5}}
	\caption{Camera pose estimation on the unordered view of ZJU-RS dataset. We evaluate the translation error (m) and rotation error ($^\circ$). For each metric, the best in \textbf{bold}.}
	\label{tab: pose estimation accuracy supplementary}
	\begin{center}
	\setlength\tabcolsep{9pt}
    \renewcommand\arraystretch{1.0}
	\setlength{\belowcaptionskip}{0pt}
	\scriptsize
 	\vspace{-2.0em}
        \small
	\resizebox{\linewidth}{!}{
		\begin{tabular}{c|cc|cc|cc|cc|cc}
			\toprule
			 &\multicolumn{2}{|c}{BARF}& \multicolumn{2}{|c}{DiffSfM}&\multicolumn{2}{|c}{Tri-MipRF-BA} &  \multicolumn{2}{|c}{USB-NeRF-RE}&  \multicolumn{2}{|c}{URS-NeRF}\\
		  &Trans& Rot&Trans& Rot &Trans& Rot&Trans& Rot & Trans & Rot \\
			\midrule
		 D0  &0.047&8.307&0.010 &\bf 1.393 &0.012 &2.104 &0.147&20.80  &\bf 0.008 &2.663   \\
      D2 &0.064 &2.999&0.015& \bf 1.075 &0.010 &1.916 &0.101&10.99  &\bf 0.007 &3.081   \\
		 D3& 0.045  &  11.258   &\bf 0.007&1.398 &0.009 &3.131 &0.195&28.91  &\bf 0.007 &\bf 1.334   \\
   	 D8  &0.021&5.058 &0.027& \bf 2.274& 0.033 & 2.264 &0.112 & 7.709  & \bf 0.014 & 2.668   \\
          C5 &0.023&5.286 & 0.010&1.787 &0.013 & 5.866 & 0.079& 11.134  & \bf 0.007 & \bf 1.694   \\
          C11 & 0.033& 4.047 &0.010&2.032 &0.009 & 1.987 & 0.081& 5.725  & \bf 0.007 &\bf 1.000  \\
        \specialrule{0.08em}{1pt}{1pt}
		\end{tabular}
		}
	\vspace{-2em}
	\end{center}
\end{table}

\section{Limitation Discussion}
As mentioned in the main paper,  introducing the estimated parameters $\mathbf{v}$ and $\bm \omega$ which are independent of the camera poses increases the degree-of-freedom of the model compared to the interpolation methods. Therefore, the accuracy of our method on sequential data, particularly for intense camera motion, is inferior to the interpolation method used in USB-NeRF. However, the flexibility and generalizability of our method are significant advantages in practical applications. Therefore, depending on the specific application scenarios, one can flexibly choose between these two methods.

\section{Video Comparison}
To further demonstrate the advantage of our method, we also present a supplementary video that demonstrates the ability of our method to recover high-quality global shutter images from the rolling shutter training images and generate images with different degrees of RS effects using the estimated velocity and angular velocity. 

\clearpage
%
%
\bibliographystyle{splncs04}
\bibliography{main}
\end{document}